\documentclass{INTERSPEECH2023}

\usepackage{amsmath,graphicx, url}
\usepackage{makecell}
\usepackage{multirow}
\usepackage{arydshln}

\DeclareMathOperator*{\argmax}{arg\,max}

\interspeechcameraready

\title{Learning Cross-lingual Mappings for Data Augmentation to Improve Low-Resource Speech Recognition}
\name{Muhammad Umar Farooq, Thomas Hain \thanks{This work was partly supported by LivePerson Inc. at the Liveperson Research Centre.}}
\address{Speech and Hearing Research Group, University of Sheffield, UK.}
\email{\{mufarooq1, t.hain\}@sheffield.ac.uk}

\begin{document}
\ninept

\maketitle
 
\begin{abstract}
Exploiting cross-lingual resources is an effective way to compensate for data scarcity of low resource languages. Recently, a novel multilingual model fusion technique has been proposed where a model is trained to learn cross-lingual acoustic-phonetic similarities as a mapping function. However, handcrafted lexicons have been used to train hybrid DNN-HMM ASR systems. To remove this dependency, we extend the concept of learnable cross-lingual mappings for end-to-end speech recognition. Furthermore, mapping models are employed to transliterate the source languages to the target language without using parallel data. Finally, the source audio and its transliteration is used for data augmentation to retrain the target language ASR. The results show that any source language ASR model can be used for a low-resource target language recognition followed by proposed mapping model. Furthermore, data augmentation results in a relative gain up to 5\% over baseline monolingual model.

\end{abstract}

\section{Introduction}
\label{sec:intro}

End-to-end (e2e) acoustic modelling techniques require a lot of training data for reliable parameters estimation. However, more than half of the world's population speak only 23 languages out of more than 7000 languages being spoken across the globe \cite{ethnologue}. Thus only a few languages have sufficient data resources, and a lot of languages are still under resourced to build an ASR system. For such languages, multilingual speech recognition systems have stolen the lime light over the past decade \cite{abate20,tachbelie20,martin16,besacier14,imseng14,vu13_interspeech} which have been used for feature extraction \cite{frantisek14,karel12,arnab13} or directly for transfer learning \cite{tong18,huang13}.

Data augmentation is another approach to increase the training data of a low-resource language. Commonly used data augmentation technique includes extending training data by making perturbed copies either by adding noise \cite{deepspeech,gales09}, varying speed and tempo of original speech \cite{ko15}, vocal tract length perturbation (VTLP) \cite{jaitly13, cui15}, SpecAugment \cite{park19} and combinations of these methods \cite{ragni14}. All these techniques are based on audio data augmentation.

In the recent past, a few studies have been done to augment data by processing text rather than speech \cite{emond18, datta20, thomas20}. Transcripts from different languages have been transliterated to Latin script to train a multilingual system \cite{datta20}. However, it requires paired data (a word in original script and its transliteration in Latin) for each language. Thomas \textit{et al.} \cite{thomas20} have proposed to transliterate a source language data to the target language without using parallel data. Source language audio data has been decoded using the target language ASR to transliterate them into target language which is then used as augmented data to retrain target language ASR.
Though this is a novel idea, an out of domain ASR has no knowledge of input language and thus is not expected to generate a good transliteration.

Recently, we have proposed a technique to learn cross-lingual acoustic-phonetic similarities on phoneme level \cite{farooq22a} which has been used for multilingual and cross-lingual acoustic model fusion \cite{farooq22b}. A model is trained to learn mappings from a source language ASR output posterior distributions to that of the target language ASR.
The study has been based on an underlying assumption that these mapping models can learn some language-related relations between phonemic posterior distributions. Though the study proves the concept, the work has been done on phoneme level using DNN-HMM hybrid systems and handcrafted lexicons for each language. In this work, we extend the previous work for cross-lingual e2e speech recognition systems. Then the ASR systems of source languages followed by a source-target mapping model for each source-target pair is used to transliterate source data into the target language script. Though both the components are trained on task specific data and are expected to generate better output labels, transliteration of a source language audio data into the target language is still unintelligible especially for unrelated languages and thus called \textit{ciphered} data. So, the key contribution of this work is two-fold;
\begin{itemize}
    \item it extends the concept of learning cross-lingual mappings for e2e speech recognition systems and
    \item generates ciphered text for a target language data augmentation using source languages ASR and $<$\textit{source-target}$>$ mapping models.
\end{itemize}

Exploiting mapping models for cross-lingual speech recognition shows that using a source language ASR for a target language gives comparable results. These mapping models are trained on limited data, and using a source language ASR followed by a mapping model enables us to exploit cross-lingual ASR to recognise the target language speech data. Furthermore, the proposed data transliteration and augmentation techniques yield up to 5\% and 28.5\% relative improvement in character error rate (CER) when compared with monolingual and multilingual ASR systems respectively.



\section{Mapping models}
\label{sec:mns}

 Let $M_{A}$ and $M_{S_{i}}$ be the monolingual acoustic models of the target and $i^{th}$ source language respectively, a mapping model $N_{S_{i}A}$ is trained to translate posteriors $P_{S_{i}}$ of dimension $d_{S_{i}}$ from $M_{S_{i}}$ to the posteriors $P_{S_{i}A}$ of dimension $d_{A}$ where $d_{A}$ is the dimension of posteriors from $M_{A}$. Given a set of observations \(X=\{x_{1},x_{2},\dotsc,x_{T}\}\) of the target language, posterior distributions (\(P^{Z}=\{p_{1},p_{2},\dotsc,p_{T}\}\) where \(Z \in \{A,S_{i}\} \)) are attained from the target and the $i^{th}$ source acoustic models. A mapping model is trained using KL divergence loss to map posteriors from $i^{th}$ source acoustic model ($P^{S_{i}}$) to the target language posteriors ($P^{S_{i}A}$). The loss function is given as;
\begin{equation}
\label{eq:kl}
\mathcal{L}_{S_{i}A}(\theta)=\sum_{n=1}^{B} p^{A}_{n} \cdot (\log p^{A}_{n}- \log p^{S_{i}A}_{n})
\end{equation}
where $B$ is the number of frames in one batch for training a mapping model $N_{S_{i}A}$ to map posteriors from $i^{th}$ source language to the target language.

\begin{figure}[t]
    \centering
    \includegraphics[width=\linewidth]{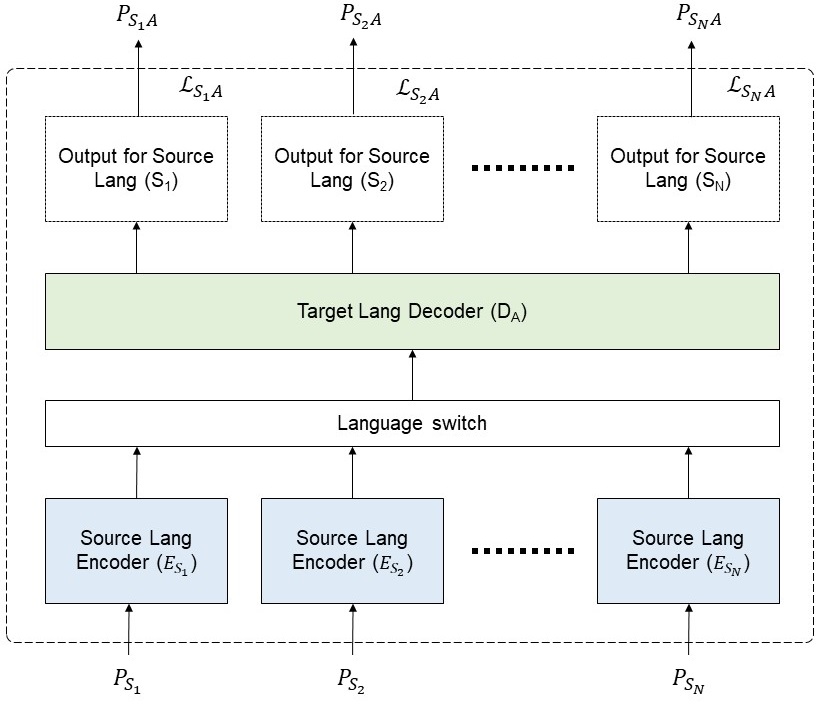}
    \caption{Architecture of the MESD mapping model}
    \label{fig:archi}
\end{figure}

Mapping models in the previous work \cite{farooq22b} have been trained on frame level without considering the contextual information but connected speech is a continuous signal which poses co-articulation and temporal smearing. Furthermore, a separate model has been trained for each source-target language pair rising a requirement of $N(N-1)$ mapping models. So, the architecture of mapping model is modified in this work to a sequence-to-sequence model with Multi Encoder Single Decoder (MESD) architecture. Thus, it incorporates contextual information and reduces the required number of mapping models to just $N$. The architecture of MESD is shown in the Figure \ref{fig:archi}.


 During the training of MESD model, outputs from all the source acoustic models for a given utterance $u$ are fed to source-language dependent encoders successively. Embeddings from the final layer of the encoders are then passed to a single target-language dependent decoder. Loss is calculated as mean of the losses for all encoder-decoder pairs.

\begin{equation}
\label{eq:kl2}
\mathcal{L}_{A}(\theta)=\sum_{K} {w_{k} \cdot \mathcal{L}_{S_{k}A}}
\end{equation}

where $K$ is the number of total source languages ($N-1$),  $w_{k}=\frac{1}{K}$ in the case of mean average and $\mathcal{L}_{S_{k}A}$ is given in Equation \ref{eq:kl} which is still frame based. It allows mapping models training to converge in low-resource setting as a small amount of data provides millions of examples. However, this causes unbalanced training across languages as mean value can be continuously decreasing when loss for one of the languages is decreasing monotonically but increasing in same fashion for the other one. This can cause model to learn mappings for one language way better than the other. To cope with this issue, a dynamic weighting scheme is applied to weight the losses for each encoder-decoder loss. For the experimentation here, rank sum weighting \cite{Roszkowska2013} is used to assign the weights. In this scheme, weights are assigned based on their normalised ranks. So, $w$ in Equation \ref{eq:kl2} now becomes

\begin{equation}
\label{eq:kl3}
w_{r}=\frac{2(K+1-r)}{K(K+1)}
\end{equation}
where $r$ is rank of the language when the languages are sorted on decreasing values of their losses. It restricts model from biasing towards a specific language or a group of languages.

Though a mapping model contains multiple encoders, any encoder can be used with decoder during decoding and MESD does not require data stream from all the encoders for a given utterance. It implies that mappings can be obtained having input even from only one source language at a time. Training of these mapping models allows to use any source language ASR for decoding the data of a target language followed by the source-target mapping model. 

\section{Ciphering text}
\label{sec:cipher}
In the previous work \cite{thomas20}, target language ASR has been used for transliteration of source language audio data for data augmentation and retraining of target language ASR. However, an ASR does not have any source language information and is not expected to generate a rationale transliterated transcriptions.

In this work on the contrary, source language audio data is decoded using in-domain ASR ($M_{S_{i}}$) and then the output posterior distributions ($P^{S_{i}}$) are transformed to the target language posterior distributions ($P^{S_{i}A}$) using the source-target mapping model ($N_{S_{i}A}$).
Mapped posteriors from the mapping models are then used to generate transliterated transcriptions (alternatively referred as \textit{ciphered} text or transcriptions) using greedy decoding. Though the transliterations still might not be exact transliterations (thus called \textit{ciphered} text), both the components involved in the process are trained using the task-specific data and are expected to perform better.

\begin{figure}[]
    \centering
    \includegraphics[width=0.7\linewidth]{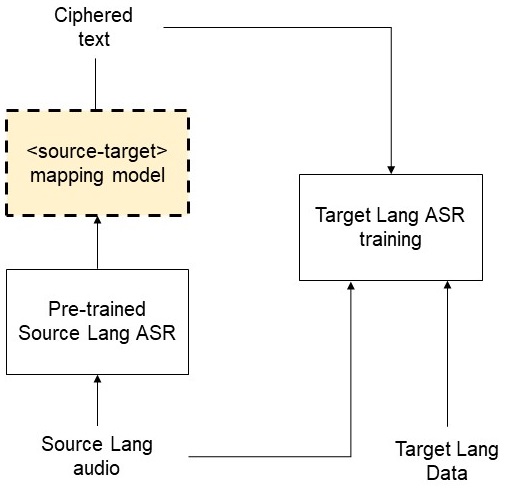}
    \caption{Flow of generating data for augmentation and retraining of target language ASR}
    \label{fig:dataug}
\end{figure}

Source language audios and their ciphered transcriptions are then used as augmented data for retraining of the target language ASR. The flow is shown in the Figure \ref{fig:dataug}.

\section{Experimental setup}
\label{sec:es}

\subsection{Data set}
\label{sec:dataset}

\begin{table}[b]
\centering
\caption{Details of BABEL data sets used for the experimentation}
\label{tab:data}
 \begin{tabular}{l|cc|cc}
\hline
\hline
\multirow{2}{4em}{Lang}&\multicolumn{2}{c|}{Train}&\multicolumn{2}{c}{Eval}\\
\cline{2-3}\cline{4-5}
&\# hours&\# spks&\# hours&\# spks\\
\hline
Tamil \textit{(tam)}&59.11&372&7.8&61\\
Telugu \textit{(tel)}&32.94&243&4.97&60\\
Cebuano \textit{(ceb)}&37.44&239&6.59&60\\
Javanese \textit{(jav)}&41.15&242&7.96&60\\
\hline
\hline
\end{tabular}
\end{table}

As this work extends the previous work, experiments here are done on same data set as used in \cite{farooq22b}. Full Language Packs (FLP) of four low-resource languages from IARPA BABEL speech corpus \cite{babel} (Tamil (\textit{tam}), Telugu (\textit{tel}), Cebuano (\textit{ceb}) and Javanese (\textit{jav})) are used for baseline ASR training and evaluation. BABEL data set mostly consist of conversational telephone speech with real-time background noises and is quite challenging because of conversation styles, limited bandwidth, environment conditions and channel. All the utterances without any speech are discarded. The details of the data sets are tabulated in Table \ref{tab:data}.

For training of the mapping models, a subset of 30 hours is randomly selected from each language pack. This data is further split into 29 hours of train set and 1 hour of dev set.

\subsection{Speech recognition systems}
\label{sec:asrs}
Hybrid CTC/attention architecture \cite{kim17} is used to train all speech recognition models which consists of three modules that are; a shared encoder, an attention decoder and a CTC module.  The training process jointly optimises the weighted sum of CTC and attention model.
\begin{equation}
\label{eq:asrloss}
\mathcal{L}_{ASR}(\theta)=\alpha \mathcal{L}_{CTC}+(1-\alpha)\mathcal{L}_{att}
\end{equation}
The input to the model is 40 filterbanks and the output of the model is the byte-pair encoded (BPE) tokens. Monolingual ASRs are trained for 100 BPE tokens for each language while the output of multilingual ASR is 400 tokens. SentencePiece library \cite{sentencepiece} is used for tokenisation.
During decoding, the final prediction is made based on a weighted sum of log probabilities from both the CTC and attention components. Given a speech input $X$, the final prediction $\hat{Y}$ is given by;
\begin{equation}
\label{eq:decode}
\hat{Y} = \argmax_{Y \in \mathcal{Y}} \{\lambda \log P_{CTC} (Y|X) + (1-\lambda)\log P_{att}(Y|X)\}
\end{equation}
where $\lambda$ is a hyper-parameter. The values of $\alpha$ and $\lambda$ are kept same for all ASR systems. SpeechBrain toolkit \cite{speechbrain} is used for training of all ASR systems.

\subsection{Mapping models}
\label{sec:mapnet}

A multi encoder single decoder model is trained for each target language. In an MESD model, there are three encoders and only one attention decoder. 
Each encoder and single decoder consists of one bidirectional RNN layer. For each target language, mapping model size is only 2.59 million parameters.

\subsection{Performance metric}
\label{sec:pm}
Accuracy of a mapping model is measured as the ratio of number of correctly mapped frames to the total number of frames as given in Equation \ref{eq:acc}. Correctly mapped frames are defined as the frames where the values of $\argmax(mapped\_posteriors)$ and $\argmax(target AM\_posteriors)$ are the same.

\begin{equation*}
    \arg \max_k(p^{A}_{t,k})==\arg \max_k(p^{S_{i}A}_{t,k})))\Rightarrow CMF\mathrel{+}+
\end{equation*}

\begin{equation}
    \label{eq:acc}
    Accuracy=\frac{CMF}{T}
\end{equation}

where $k$ is the index of classes in the output vector $p_{t}$, $CMF$ is the number of correctly mapped frames and $T$ is the total number of frames.

For downstream speech recognition task, results are reported in terms of percent character error rate.

\section{Results and Discussion}
\label{sec:rnd}

\subsection{Mapping models}
\label{sec:mnr}

Accuracies of mapping models, trained to map posterior distribution from a source language ASR to the target language ASR, are tabulated in Table \ref{tab:mapAccu}. Analysis shows that correct target class is still among top $n$ mapped classes if not the most probable one. So, the mapping models accuracy is calculated for different values of $n$ where $n$ represents the number of most probable classes. Though the accuracy increases with increasing value of $n$, rate of change is not as much as observed in case of phonemes by \cite{farooq22b} which implies that the performance of mapping model in case of phoneme based hybrid DNN-HMM systems has been better than that for e2e systems. Since the mapping models are trained using posterior distributions of ASR outputs, one potential reason could be the detrimental affect of speech recognition systems on the training of mapping models. However, the joint analysis of amount of training data (Table \ref{tab:data}), performance of monolingual speech recognition systems (Table \ref{tab:baseline}) and performance of mapping models (Table \ref{tab:mapAccu}) rules out this reason. 

Amount of mapping model training data is same for all the languages but the mappings for \textit{ceb} and \textit{jav} target language is better than \textit{tam} and \textit{tel}. Even for \textit{ceb} and \textit{jav} target languages, accuracy of mappings from \textit{tel} source language is very low in comparison to other source languages. The investigation reveals that as the number of BPE tokens are restricted to 100 for all the languages, \textit{ceb} and \textit{jav} having only 19 and 26 characters respectively have good context coverage in 100 BPE tokens. But the BPE tokens extracted for \textit{tel}, which have more than 52 characters, do not cover context very well. Furthermore, both \textit{ceb} and \textit{jav} are written in Latin script and thus have a full overlap of characters and are even acoustically close. While on the other hand though both \textit{tam} and \textit{tel} belong to same Dravidian family, their writing scripts are different which makes it difficult for model to learn mappings with limited number of BPE tokens.

\begin{table}[b]
    \centering
    \caption{Accuracy of the mapping models considering top n mapped classes}
    \label{tab:mapAccu}
    \begin{tabular}{llcccc}
    \hline \hline
        \multirow{2}{4em}{Target Lang}&\multirow{2}{4em}{Source Lang}&\multicolumn{4}{c}{\textit{Mapping model} accuracy}\\
        \cline{3-6}
        &&\textit{n=1}&\textit{n=2}&\textit{n=5}&\textit{n=10}\\
    \hline
        \multirow{3}{4em}{tam}&tel&47.46&54.58&66.31&77.06\\
        &ceb&45.98&52.88&64.25&74.65\\
        &jav&46.97&54.02&65.63&76.26\\
    \hline
        \multirow{3}{4em}{tel}&tam&48.88&56.20&67.80&78.28\\
        &ceb&46.22&53.27&64.97&75.96\\
        &jav&47.40&54.78&66.76&77.54\\
    \hline
        \multirow{3}{4em}{ceb}&tam&60.53&66.32&74.79&82.31\\
        &tel&48.32&51.43&56.49&62.53\\
        &jav&65.04&71.39&80.06&86.58\\
    \hline
        \multirow{3}{4em}{jav}&tam&62.24&68.40&77.00&83.76\\
        &tel&54.64&57.92&62.29&67.69\\
        &ceb&65.51&71.85&80.30&86.65\\
    
    \hline \hline
    \end{tabular}
    
\end{table}


\subsection{Ciphering text}
\label{sec:ctr}

\begin{figure}
    \centering
    \includegraphics[width=\linewidth]{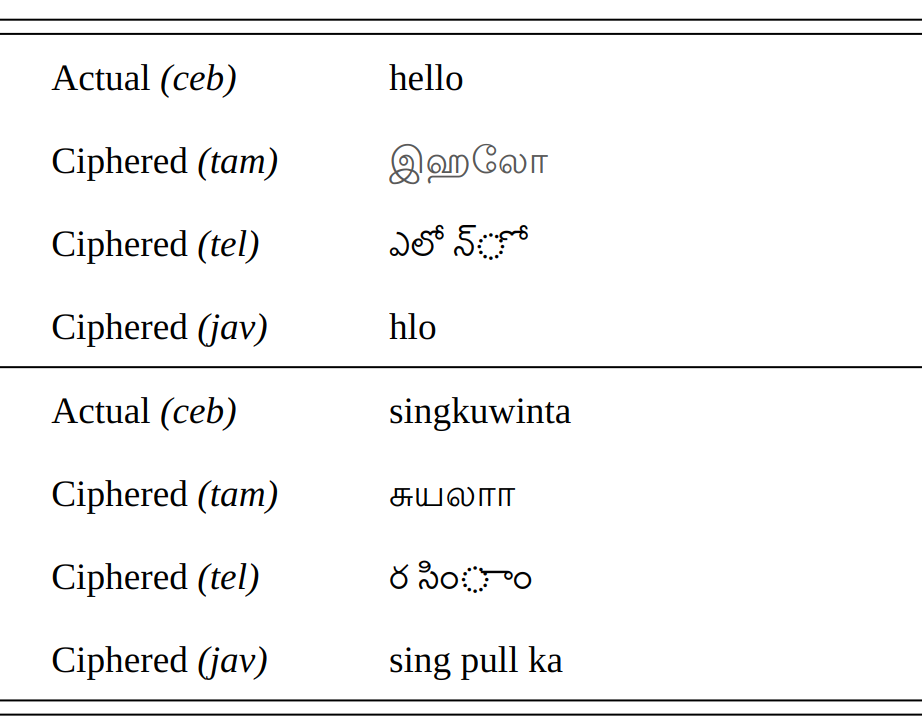}
    \caption{Examples of ciphered transcriptions}
    \label{fig:cipexamps}
\end{figure}

For a given target language, audio data of all the source languages is decoded using language dependent ASR systems and the output posterior distributions are then mapped to target language distributions using the mapping models $N_{S_{i}A}$. Greedy decoding is carried out on these output posterior distributions to generate ciphered transcriptions for the target language. Language model (LM) is not integrated at this stage to avoid LM affect on transliterations. As this stage solely depends on mapping models, the quality of ciphered text depends on mapping models accuracy for $n=1$. The analysis of ciphered transcriptions shows that the transliteration is fairly good for shorter utterances but gets worse for longer utterances. A few examples of ciphered transcriptions are shown in Figure \ref{fig:cipexamps}.

\subsection{ASR}
\begin{table}[t]
\centering
\caption{ASR performance in terms of \%CER}
\label{tab:baseline}
 \begin{tabular}{lcccc}
\hline
\hline
Lang&tam&tel&ceb&jav\\
\hline
\textit{mono}&44.6&58.24&39.40&42.42 \\
 + LM &39.25&52.68&31.25&32.11 \\
\textit{multi}&41.15&54.38&38.91&42.65\\
\hdashline
\textit{augAll}&41.90&56.10&32.30&32.86\\
\textit{augTwo}&\textbf{38.83}&\textbf{52.06}&\textbf{29.94}&\textbf{30.47}\\

\hline
\end{tabular}
\end{table}

Monolingual systems (\textit{mono}) are the language dependent acoustic and language models which are trained on target language specific data sets. The train sets of all the languages are then mixed to train a multilingual system (\textit{multi}). Language model for a multilingual system is also trained using mix corpora of individual languages. The results of speech recognition systems are shown in Table \ref{tab:baseline}. The first row contains the monolingual ASR result without using LM for a later comparison while rest of the results are ASR decoding with LM.

For a given target language test set, speech recognition results are also computed on top of mapping models after decoding target language data using source language acoustic models. Greedy decoding is applied on mapped posteriors and the results are shown in Table \ref{tab:cross}. CER on diagonal is the same as the first row of Table \ref{tab:baseline}. Though these results are from source language ASR followed by a source-target mapping model and does not use language dependent ASR, it preforms better than monolingual ASR in case of \textit{jav}. Results are comparable for other languages but fairly depend on mapping models performance. It is evident from these results that a source language acoustic model can be used for decoding of a target language followed by a mapping model trained on limited amount of data. 

\begin{table}[b]
    \centering
    \caption{Cross-lingual ASR performance in terms of \%CER}
    \label{tab:cross}
    \begin{tabular}{lcccc}
    \hline \hline
        \multirow{2}{4em}{Target Lang}&\multicolumn{4}{c}{\textit{Source Languages} accuracy}\\
        \cline{2-5}
        &\textit{tam}&\textit{tel}&\textit{ceb}&\textit{jav}\\
        \hline
        \textit{tam}&\textit{44.60}&49.34&49.21&49.03\\
        \textit{tel}&63.19&\textit{58.24}&64.33&63.65\\
        \textit{ceb}&48.10&65.31&\textit{39.40}&40.94\\
        \textit{jav}&46.72&56.92&40.88&\textit{42.42}\\
    \hline \hline
    \end{tabular}
    
\end{table}

\subsubsection{Data augmentation}
Ciphered transcriptions are generated from all the source languages for a target language using mapping models as described in Section \ref{sec:cipher}. Then the audio data of source languages and the ciphered transcriptions are used together as augmented data for retraining of target language ASR (\textit{augAll}). As described earlier, the quality of ciphered transcriptions depends on performance of mapping models, using ciphered transcriptions data augmentation from all the source languages include very low quality transcriptions and have detrimental effect on retraining of target language ASR. So, the augmentation is then restricted to use ciphered data from only closest language (\textit{augTwo}).
For a target language, the source language with highest mapping model accuracy is chosen as the closest language.
By augmenting this data for retraining of a target language, an relative gain of up to 5\% is achieved in terms of CER (\textit{augTwo}).


\section{Conclusion}
In this work, the technique of mapping models is extended for e2e speech recognition systems. For a given target language, a mapping model is trained on limited amount of data to transform output posterior distributions from a source language ASR model to that of the target language. A source language ASR followed by a mapping model is then used for cross-lingual speech recognition in low-resource setting. Mapping models are further exploited to transliterate data of a source language to the target language for data augmentation. Retraining of target language ASR after data augmentation results in a relative CER reduction of up to 5\% and 28.5\% in comparison to monolingual and multilingual ASR systems respectively.

\bibliographystyle{IEEEtran}
\bibliography{template}

\end{document}